\documentclass{article}

\usepackage[preprint]{neurips_2021}

\usepackage[utf8]{inputenc} 
\usepackage[T1]{fontenc}    
\usepackage{hyperref}       
\usepackage{url}            
\usepackage{booktabs}       
\usepackage{amsfonts}       
\usepackage{nicefrac}       
\usepackage{microtype}      
\usepackage{xcolor}         

\usepackage{amsmath}
\usepackage{graphicx}
\usepackage{multirow}
\usepackage{bm}

\newcommand{\etc}{\textit{etc}.}
\newcommand{\eg}{\textit{e}.\textit{g}.}
\newcommand{\etal}{\textit{et al}. }
\newcommand{\ie}{\textit{i}.\textit{e}.}

\title{MOC-GAN: Mixing Objects and Captions to Generate Realistic Images}

\author{%
  Tao Ma$^1$ \ \ \ \ \ \ \ \ \ \ \ \  Yikang Li$^{1,2}$\thanks{Corresponding author.} \\
  $^1$Shanghai AI Laboratory, $^2$SenseTime \\
  \texttt{matao@pjlab.org.cn, liyikang@pjlab.org.cn} \\
}

\begin{document}

\maketitle

\begin{abstract}
Generating images with conditional descriptions gains increasing interests in recent years. However, existing conditional inputs are suffering from either unstructured forms~(captions) or limited information and expensive labeling~(scene graphs). 
For a targeted scene, the core items, objects, are usually definite while their interactions are flexible and hard to clearly define. 
Thus, we introduce a more rational setting, generating a realistic image from the objects and captions. Under this setting, objects explicitly define the critical roles in the targeted images and captions implicitly describe their rich attributes and connections. 
Correspondingly, a MOC-GAN is proposed to mix the inputs of two modalities to generate realistic images. 
It firstly infers the implicit relations between object pairs from the captions to build a hidden-state scene graph. So a multi-layer representation containing objects, relations and captions is constructed, where the scene graph provides the structures of the scene and the caption provides the image-level guidance. Then a cascaded attentive generative network is designed to coarse-to-fine generate phrase patch by paying attention to the most relevant words in the caption. In addition, a phrase-wise DAMSM is proposed to better supervise the fine-grained phrase-patch consistency. On COCO dataset, our method outperforms the state-of-the-art methods on both Inception Score and FID while maintaining high visual quality. Extensive experiments demonstrate the unique features of our proposed method.
\end{abstract}

\section{Introduction}
Generating a realistic image as the human's expectation is a challenging but exciting task. Researchers have investigated various conditional inputs to supervise the image generation, such as a semantic map~\cite{qi2018semi,wang2018high}, a reference image~\cite{isola2017image}, a description sentence~\cite{xu2017attngan,han2017stackgan,Han17stackgan2} or a scene graph~\cite{johnson2018image,li2019pastegan}. Among these kinds of methods, the textual descriptions like sentence and the scene graph are user-friendly ones to generate the targeted images, as they provide us an easy way to depict our expectations without any requirement for artistic attainment. Humans can freely control the scene appearance by defining the objects and connecting them with pair-wise relations~(scene graphs) or even typing a free-form sentence~(image captions) to describe their overall expectations.

However, both scene graphs and sentences have their shortcomings. For the former setting, which depicts the image with objects and their pair-wise relations, it only covers pair-wise interactions but lacks the high-ordered semantics. Additionally, it is hard to describe the complicated relationships with only one label from a limited and long-tail-distributed predicate dictionary. For the caption sentences, they can provide multifarious descriptions including objects, their attributes and high-order interactions. But it is an abstract and high-level representation, which lacks the fine-grained definition of every item in the scene, making it volatile on generating complicated scenes. In addition, it is also hard to conduct fine-grained control of the generation process. Therefore, to endow the human with the ability to freely create the images, the conditional input should contain both precise and general information of the scene. 

To this end, we formulate the task as generating the image from both the objects and a high-level caption. The former determines the base items of the generated image and the latter provides a general depiction of the scene.
To build the connection between the objects and the caption, we introduce an attention-based Implicit Graph Estimator to infer the pair-wise relation semantics between objects from the captions. Then a scene graph is constructed as the backbone of the scene while the caption semantics provides the image-level guidance. 
Different from the edge embedding in \cite{johnson2018image,li2019pastegan}, which is explicitly defined with a limited predicate dictionary, our relation embedding learns a soft semantics which covers a larger interaction semantic space based on the textual descriptions. Moreover, annotating captions is more efficient than scene graph, as it is hard to exhaustively annotate the pair-wise relations and the annotations skew toward the trivial categories, like ``on'' or ``near''. 

Along with the inferred relation embeddings, we have an integrated description of the scene containing different semantic levels. 
As objects and relations contain spatial information~(bounding boxes), we could generate a phrase-wise layout map indicating the spatial arrangements of the visual cues based on the implicit graph. 
In addition,  we also prepare another graph semantic map using upsample\&conv structure from the global graph vector to encode the semantic information about the scene from the implicit graph.
Besides, to enhance the details of generated images, we propose a phrase-wise DAMSM to capture the phrase-context features and supervise the subgraph-based similarity between phrases and similar interacting sub-regions of image.
These semantic maps will define the generated image both spatially and semantically.

The contributions of this work are summarized as three-fold: 
$(i)$ We introduce a more reasonable setting of conditional image generation, which has several unique features, \eg, easily-collected annotations, fine-grained manipulation of the items, high-level control of the scene, \etc. 
$(ii)$ A MOC-GAN is proposed to mix the objects and scene captions to synthesize realistic images. An implicit graph is built by inferring the object-pair relations from the caption with the proposed Implicit Relation Estimator. A Hidden Feature Aggregator is designed to blend the features of different semantic levels and generate a high-resolution image. Additionally, a phrase-wise DAMSM is used to supervise the consistency between the implicit graph and corresponding regions.
$(iii)$ Evaluated on COCO dataset, our MOC-GAN outperforms the SOTA methods conditioned on either scene graphs or captions. Qualitative results also show our superior performance on generating images involving multiple objects. 

\section{Related Work}
\noindent \textbf{Scene Graph.}
A Scene graph is a topological representation of a scene which represents object instances as nodes and relationships between objects as edges~\cite{johnson2015imgret}.
Scene graphs have been employed in various works, \eg, image retrieval~\cite{johnson2015imgret}, image captioning evaluation~\cite{spice2016}, sentence to scene graph translation~\cite{xu2017sggenimp}, and image-guided scene graph generation~\cite{DBLP:conf/nips/2018,Li2018fnet,Li2017msdn}.
Recently, scene-graph-based generative models have shown uncommon ability on capturing the structure information between nodes. The core part of these methods is the generation of the seed feature maps that are fed into the image decoder. Johnson~\etal use a graph convolution to process the input scene graph and then compute the layout map by filling the object features into a noise map~\cite{johnson2018image}.  Li~\etal introduce an auxiliary object image crop to finely control the object appearance and use a 2D graph convolution to generate the layout map in a learnable way~\cite{li2019pastegan}.
Though capturing integral structure information, scene-graph-based methods always lack of the fine-grained details of the scene during the generation process.
In this paper, instead of using the explicitly defined relation labels, we propose to infer the pair-wise relation embeddings from the caption through a phrase-driven attention mechanism, which could better encode the interaction and spatial relations between object pairs.

\noindent \textbf{Text to Image.}
Generating images from text descriptions, though challenging, has recently drawn a lot of attention from many real-world tasks, \eg, art generation and computer-aided design. With the emergence of deep generative models, a lot of great progress has been achieved in this direction, such as Variational Auto-Encoders (VAE)~\cite{mansimov16_text2image}, optimization techniques~\cite{nguyen2016ppgn} and auto-regressive models~\cite{Reed1}.
Compared with other approaches, methods based on conditional Generative Adversarial Networks (GANs)~\cite{goodfellow2014gan} present promising results in text-to-image generation. Reed \etal first learned that both generator and discriminator conditioned on text embedding is capable of generating great images~\cite{ReedAYLSL16}. Zhang \etal increased image resolution and draw images of different sizes by stacking several GANs~\cite{han2017stackgan,Han17stackgan2}.
Although these approaches are capable of high-quality image generation on datasets of specific objects (flowers~\cite{flowers} and birds~\cite{WelinderEtal2010}), the perceptual quality of synthesis tends to decline on datasets with multiple objects and complicated scenes (COCO dataset~\cite{10.1007/978-3-319-10602-1_48}), because existing models are always conditioned on the global caption feature so that miss important structure and interaction information.
Therefore, our MOC-GAN adopts object labels to form a new implicit graph structure to maintain the spatial and interaction during the image generation.

\noindent \textbf{Attention Mechanism.}
The application of attention mechanism in vision-language multi-modality intelligence tasks has also been studied recently.
Many works~\cite{xu2017attngan,XuBKCCSZB15,DAGAN,YangHGDS15,Zhang2018SelfAttentionGA,stackcross} have been proposed to use grid attention mechanism in modeling multi-level dependencies.
The Deep Attentional Multimodal Similarity Model (DAMSM), measuring the image-text similarity at the word level, has been widely used in many generation tasks~\cite{li2019objgan,xu2017attngan,yin2019sdgan} and proved the effectiveness of supervising the hidden feature maps during generation.
In this paper, we expand this pre-trained attention mechanism as phrase-wise DAMSM for graph convolution network processing graph structure and general convolution network handling the visual data.
With this scheme, the phrase features would better align a sub-region containing corresponding object pairs in the image, as well as their interactions.

\section{Methodology}
\begin{figure*}[t]
\centering
\includegraphics[width=1.\textwidth]{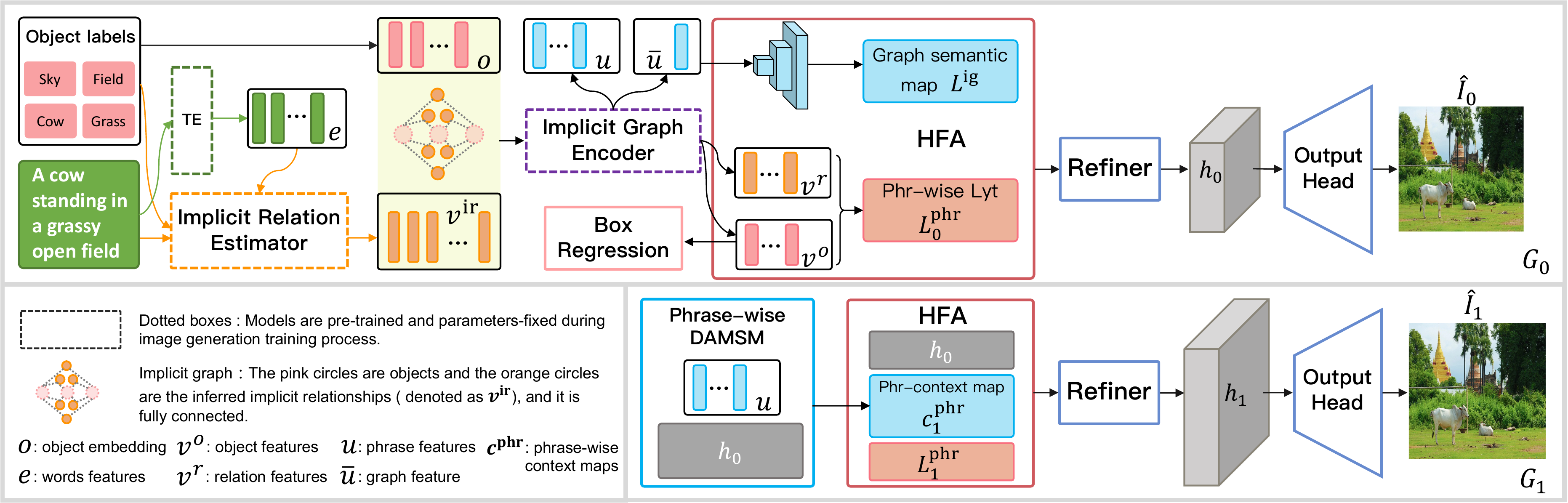}
\vspace{-6mm}
\caption{Overview of our proposed MOC-GAN. 
}
\label{fig:framework}
\vspace{-6mm}
\end{figure*}
The overall pipeline of our proposed MOC-GAN is illustrated in Figure~\ref{fig:framework}.
Given the object labels and a supplementary caption, our model could generate a realistic image, in which the specific attributes of objects and interaction relationships between objects are consistent with the caption. We first utilize a bi-LSTM based Text Encoder (TE) to process the caption as words features $e \in \mathbb{R}^{{D_w} \times T_w}$, and then introduce an Implicit Relation Estimator~(IRE) to infer the interaction relationships between any two objects by paying attention to the most relevant words in the caption.
Thus, a virtual graph could be generated depicting the objects and their pair-wise interaction relations implicitly.
This virtual graph, named implicit graph, is then directly fed into our pre-trained graph-convolution-based Implicit Graph Encoder (IGE) to get two groups of feature vectors. One is phrase features $u \in \mathbb{R}^{{D_p} \times T_p }$ and global graph feature $\overline{u} \in \mathbb{R}^{D_p} $ with abundant phrase-wise context information. The other contains the object vectors and relation vectors $v^o, v^r \in \mathbb{R}^{D_p}$ representing the category and spatial information, which are used to predict the spatial locations $b_i$ for each object through a Box Regression Network (BRN).
Afterwards, Hidden Feature Aggregator (HFA) would fuse all these features from different semantic spaces into pre-image domain to get a comprehensive semantic hidden map, which will be fed into Image Decoder (composed of Refiner $R_i$ and Output Head $H_i$) to generate images $\hat{I_i}$ with resolution doubling between consecutive steps from $64\times64$ to higher.
The generation process involves three cascaded steps, wherein two stages $(G_0, G_1)$ are shown in Figure~\ref{fig:framework} and could be formulated as
\begin{equation}
\label{generation}
\begin{split}
	h_0 &= R_0(\text{HFA}(L_0^{\text{phr}}, L^{\text{ig}})), \ \hat{I}_0 = H_0(h_0), \\
	h_i &= R_i(\text{HFA}(h_{i-1}, L_i^\text{phr}, c_i^{\text{phr}})), \ \hat{I}_i = H_i(h_i),
\end{split}
\end{equation}
where $(i)$ $L_i^{\text{phr}} = F_\text{Lyt}^{\text{phr}}(v^o, v^r, b)$ is the phrase-wise layout map by element-wisely summing an object layout map and relation layout map;
$(ii)$ $L^{\text{ig}} = F^{\text{ig}}(\overline{u})$ is the graph semantic map;
$(iii)$ $c_i^{\text{phr}} = F_{\text{attn}}^\text{phr}(u, h_{i-1})$ are phrase-wise context features from our proposed phrase-wise DAMSM.
The whole model is trained adversarially with several conditional and unconditional discriminators to enable the stability of generation process.
We will introduce the main innovations in the following sections and more specific details could be found in supplementary materials.

\subsection{Implicit Relation Estimator}
From a perspective of providing the object pairs with abundant interaction information, we propose the phrase-driven attention.
GloVe model~\cite{pennington2014glove} is utilized for word and object label representation. The advantage is that both of them get represented in the same high dimensional space, in which linguistically similar words will have similar representations, so the semantically related object pairs and words will lie close to each other and have a high similarity score.

\begin{figure}
\begin{center}
\includegraphics[width=0.95\linewidth]{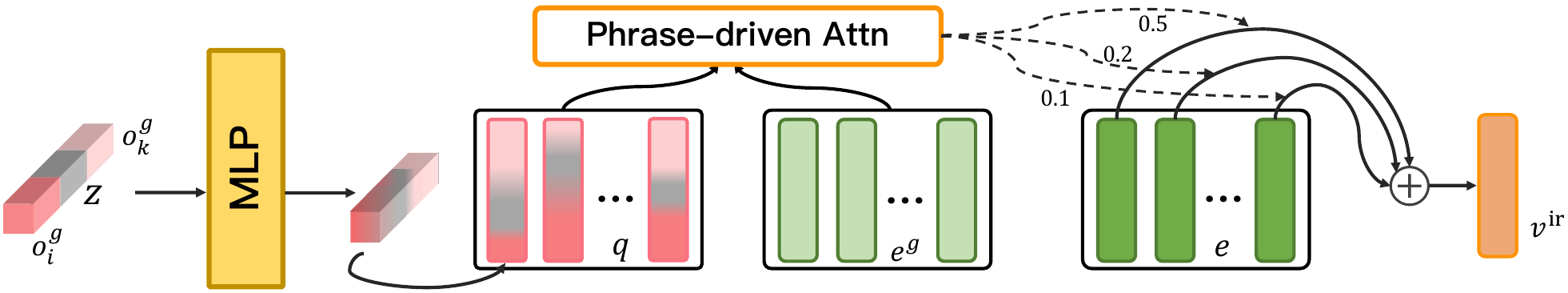}
\end{center}
\vspace{-3mm}
\caption{The proposed Implicit Relation Estimator. $o^g$ and $e^g$ are embedding of object labels and words represented by GloVe model respectively. The implicit relationship $v^\text{ir}$ is a dynamic vector attended from words embeddings $e$ output by TE.}
\label{fig:phrase-wise-attn}
\vspace{-5mm}
\end{figure}

As shown in Figure~\ref{fig:phrase-wise-attn}, we firstly concatenate the GloVe embedding $o_i^{g}$ and $o_k^{g}$ of every two objects with a noise vector $z$ sampled from a standard normal distribution to form a phrase feature. Then an MLP is utilized to convert this phrase feature to a semantic space which is used to calculate the proposed phrase-driven attention.
Specifically, the $j$-th implicit relationship vector $v_j^{\text{ir}}$ is a dynamic representation of words vectors $e$ relevant to phrase vector $q_j$, which is calculated by
\begin{equation}
\label{eq:implicit-relation}
  v_j^{\text{ir}} = \sum_{i=0}^{T_{w}-1}\beta_{j,i}^{\text{phr}} e_{i}, \; \;
  \beta_{j,i}^{\text{phr}} = \frac{\exp(s_{j,i}^{\text{phr}})}{\sum_{k=0}^{T_{p}-1}{\exp(s_{j,k}^{\text{phr}})}}, \; \;
  s_{j,i}^{\text{phr}} = (q_j)^T e_{i}^{g}.
\end{equation}
$\beta_{j,i}^{\text{phr}}$ indicates the weight the model attends to the $i$-th word when generating $j$-th phrase. $T_{w}$ and $T_{p}$ denote the number of words and phrases respectively.
Finally, we could connect the object vectors $v^o_i$ and $v^o_k$ with corresponding implicit relationship $v^\text{ir}_j$ to form a phrase of implicit graph.
There will be $T_p = n(n-1)$ implicit relationships all together, $n$ is the number of objects.

\subsection{Phrase-wise DAMSM}
Phrase-wise DAMSM is a module to better align the phrase features of the implicit graph and sub-regions of the image. This pre-training process involves three neural networks: IRE to form the implicit graph, IGE to extract local phrase features and global graph features, and Image Encoder for extracting image visual features. Once this pre-training is finished, they are all parameters-fixed.

\noindent \textbf{Implicit Graph Encoder.}
Our IGE, composed of several graph convolution layers and MLPs, is utilized to process the implicit graph. Each graph convolution layer performs similar functions: taking as input an implicit graph with vectors at each node~(object) and edge~(implicit relation), and computing new vectors $v^o, v^r$ for each node and edge, which could maintain more information because they are a function of neighborhoods of their corresponding inputs.
Subsequently, an MLP takes the concatenation of object and relation feature vectors and outputs all the local phrase features $u \in \mathbb{R} ^ {D_p \times T_p}$. Then, we fuse these phrases together through average calculation, they are processed by another MLP to get the global implicit graph vector $\overline{u} \in \mathbb{R} ^ {D_p}$ which includes the fundamental representation of the whole scene.

\noindent \textbf{Image Encoder.}
Inception-v3~\cite{sze2016incept3} model has been widely used as a backbone to extract visual features, so we keep the same settings in ~\cite{xu2017attngan} to extract two visual features: $f \in \mathbb{R} ^{D_p \times 289}$ are the visual feature vectors for all the sub-regions, and $\overline{f} \in \mathbb{R} ^{D_p}$ is the global vector for the whole image.

\noindent \textbf{Pre-training.}
With obtaining these two groups of features ($u$ and $\overline{u}$, $f$ and $\overline{f}$), we could compute the similarity between phrases and image sub-regions, as well as the similarity between the implicit graph and whole image, in the form of cosine distance between any phrase vector and grid feature vector. And this similarity serves as the objective function to optimize the IGE to learn the phrase-image consistency. The pretraining strategy of \cite{xu2017attngan} is followed. Please refer to the supplementary material for more details.

\subsection{Hidden Feature Aggregator}
The HFA plays a role of fusing features from different semantic spaces together into pre-image domain, which can be considered as the blueprint of image in feature space. To better fuse all these semantic information from upstream, we formulate three feature maps with different semantic meanings: spatial layout map $L^{\text{phr}}$, context feature map $c_i^{\text{phr}}$, graph semantic map $L^{\text{ig}}$.

\noindent \textbf{Spatial Layout Composition.}
For a couple of objects $o_i$ and $o_k$, we first fill the region of a zero-value layout $L \in \mathbb{R}^{H \times W \times D_P}$ with object vectors $v^o_i, v^o_k$ output by IGE in their respective predicted bounding boxes.
Actually, considering the interaction between objects, it's intuitive that the rest region, at least the region between $o_i$ and $o_k$ shouldn't be zeros.
Hence, we reproduce another feature map by filling the relation feature $v^r_j \in \mathbb{R}^{D_p} $ into the rest region of box-union of $b_i$ and $b_k$.
Then we element-wisely sum these feature maps of all phrases $(v^o_i, v^r_j, v^o_k)$ coming from one image to get the phrase-wise layout map $L^{\text{phr}}$. If there are multiple boxes of phrases covering the same pixels, we simply use max-operation to decide whose feature value should be retained here.

\noindent \textbf{Phrase-wise Context.}
After the primary stage, the phrase feature vectors $u$ are used to compute a phrase-wise context feature map $c_i^{\text{phr}} = F_{\text{attn}}^\text{phr}(u, h_{i-1})$ through the proposed phrase-wise DAMSM. Specifically, for $j$-th coloumn of $h_{i-1}$ , its phrase-wise context vector is a dynamic representation of phrase vectors relevant to sub-region $j$ as following,
\begin{equation}
\label{eq:phrase-attn}
  c_{i, j}^{\text{phr}} = \sum_{m=0}^{T_{p}-1}\beta_{j,m}^{\text{phr}} u_{m},
  \ 
  \text{where}
  \ 
  \beta_{j,m}^{\text{phr}} = \frac{\exp(s_{j,m}^{\text{phr}})}{\sum_{k=0}^{T_{p}-1}{\exp(s_{j,k}^{\text{phr}})}}.
\end{equation}
Here, $s_{j,m}^{\text{phr}} = (h_{i-1, j})^T u_{m}$, and $\beta_{j,m}^{\text{phr}}$ indicates the weight the model attends to the $m$-th phrase when generating $j$-th sub-region.

\noindent \textbf{Graph Semantic Map.}
In the contrast, because graph feature vector $\overline{u}$ is a global representation of an image, so we are unable to use the explicit location information. We directly enlarge the resolution of these feature vectors equaling to the image size by utilizing several upsample layers followed by convolutional layers, to form graph semantic map $L^{\text{ig}}$.

Obviously, $L^{\text{phr}}$ determines the numbers and categories of objects in this image and specifies the spatial location information. At the high semantic level, the phrase-wise context feature $c^\text{phr}$ provides the attended fine-grained details, and $L^{\text{ig}}$  defines the global semantic information and the overall description of implicit graph.
Eventually, the concatenation of $L^{\text{phr}}$, $L^{\text{ig}}$ and $c^\text{phr}$ is fed into two res-blocks for better fusion and get a comprehensive semantic map.

\subsection{Discriminators}
To guarantee the quality of the output images, and make the objects recognizable and the interaction relationships between objects reasonable, we adopt a series of patch-wise, object-wise and phrase-wise discriminators $D^{\text{pat}}, D^{\text{obj}}, D^{\text{phr}}$ to train the cascaded generative network adversarially. All the discriminator are optimized by maximizing following objective
\begin{equation}
\begin{split}
	\mathcal{L}_{GAN} = \mathop{\mathbb{E}} \limits_{x \sim p_{\mathrm{real}}} \log D(\cdot) + \mathop{\mathbb{E}} \limits_{x \sim p_{\mathrm{fake}}} \log D(1-D(\cdot)) \ .
\end{split}
\end{equation}

\noindent \textbf{Patch-wise Discriminators.}
The patch-wise discriminator promotes the image to be realistic through determining this image as real or fake (unconditional), and whether this image is consistent with the implicit graph or not (implicit graph conditional).
Given an image $x$ and corresponding implicit graph feature $\overline{u}$, these two possibilities can be respectively written as
\begin{equation}
\begin{split}
	p^{\text{pat}}_{\text{unc}} = D_{\text{unc}} ^ {\text{pat}} (x), \; 
	p^{\text{pat}}_{\text{ig\_con}} = D_{\text{ig\_con}} ^ {\text{pat}} (x, \overline{u}).
\end{split}
\end{equation}

\noindent \textbf{Object-wise Discriminators.}
Given a cropped and resized object crops, the object-wise discriminator could encourage each object in this image appear realistic and clear. Additionally, an auxiliary object classifier is added to predict the category of the object to ensure that the objects are recognizable.
The object-wise discriminator takes as input the image $x$ and bounding boxes $b_i, .., b_t$ in this image to help supervise the quality of objects. And they can be indicated as following:
\begin{equation}
\begin{split}
	p^{\text{obj}} = D^{\text{obj}} (x, b_i), \;
	p_{\text{ac}}^{\text{obj}} = D_{\text{ac}}^{\text{obj}} (x, b_i).
\end{split}
\end{equation}

\noindent \textbf{Phrase-wise Discriminator.}
We propose a novel phrase-wise discriminator to supervise the relative implicit relationship between a couple of objects in the form of $\langle subject, predicate, object \rangle $ phrase.
Given an image $x$, ground-truth bounding boxes of objects, the phrase-wise discriminator can be determined as the following:
\begin{equation}
\begin{split}
	p_{\text{unc}} ^{\text{phr}} = D_{\text{unc}}^{\text{phr}} (h^{\text{sub}}, h^{\text{pred}}, h^{\text{obj}}), \ 
	p_{\text{con}} ^{\text{phr}} = D_{\text{con}}^{\text{phr}} (h^{\text{sub}}, v^{\text{ir}}, h^{\text{obj}}).
\end{split}
\end{equation}
where $h^{\text{sub}} = \text{VGG}(x, b^{\text{sub}}), h^{\text{obj}} = \text{VGG}(x, b^{\text{obj}})$, and $h^{\text{pred}} = \text{VGG}(x, b^{\text{sub}}, b^{\text{obj}})$.

We first extract the feature of each object through a parameter-fixed VGG19 Network~\cite{2015vgg} pre-trained on ImageNet~\cite{fei2015imagenet}, as well as the \textit{predicate} feature whose box is defined as the union of \textit{subject} and \textit{object} bounding boxes.
Then the concatenation will be fed in the next classifier to determine whether this phrase feature is realistic and reasonable (unconditional). 
Another part is conditioned on the implicit relationship feature $v^{\text{ir}}$ rather than $h_{pred}$ to determine whether these two objects are correlated with the relations attended from words features (conditional).

\subsection{Training}
Our model needs several preparative models before the training of image generation, the first process is to train TE through traditional DAMSM. Then, the parameters of TE are fixed to provide words features as source materials for encoding the implicit relationships.
The second pre-training process is for phrase-wise DAMSM, including IRE and IGE, to connect the phrase features and corresponding sub-regions, the global graph features and the whole image. After the pre-training processes, the weights of TE, IRE and IGE are all fixed in the following training process.

The generator and discriminators are trained end-to-end in an adversarial manner. The generator is trained to minimize the weighted sum of the following eight losses:

\noindent \textbf{Image Adversarial Loss}
$\mathcal{L} {_{GAN} ^{img}}$ from $D{_{\text{unc}}^{\text{pat}}}$, $D{_{\text{cap\_con}}^{\text{pat}}}$ and $D_{\text{ig\_con}}^{\text{pat}}$ encourage generated image patches appear realistic, consistent with corresponding textual description and implicit graph respectively.

\noindent \textbf{Pixel Loss}
$\mathcal{L}{_1 ^{img}} = \lVert I - \hat I \rVert {_1} $ penalizes the $L1$ differences between the ground-truth image $I$ and the generated image $\hat I$, which benefits stable convergence of training.

\noindent \textbf{Perceptual Loss}
\cite{Johnson2016Perceptual} $\mathcal{L} {_P ^{img}}$ penalizes the $L1$ difference in the feature space output by VGG19 between the ground-truth image $I$ and the generated image $\hat I$.

\noindent \textbf{Object Adversarial Loss}
$\mathcal{L} {_{GAN} ^ {obj}}$ from $D^{\text{obj}}$ encourages each generated object to look realistic.

\noindent \textbf{Auxiliary Classifier Loss}
$\mathcal{L} {_{AC} ^{obj}}$ from $D_{\text{ac}}^{\text{obj}}$ encourages that each generated object to be recognizable and could be classified as the correct category.

\noindent \textbf{Phrase Adversarial Loss}
$\mathcal{L} {_{GAN} ^{phr}}$ from $D_{\text{unc}}^{\text{phr}}$ and $D_{\text{con}}^{\text{phr}}$ helps supervise the relation between any two objects in the generated images to be real and consistent with the implicit relationship.

\noindent \textbf{Phrase-wise DAMSM Loss}
$\mathcal{L}_{DAMSM}^{phr} $ computes a fine-grained loss to measure the similarity between image and implicit graph at the semantic level.

\noindent \textbf{Box Regression Loss}
$\mathcal{L} _{box} = \sum {_{i=1} ^n} \lVert b_i - \hat{b}_i \rVert$ penalizes the $L1$ difference between ground-truth and predicted bounding boxes.

Therefore, the final loss function of our model is defined as:
\begin{equation}
\small
\begin{split}
	\mathcal{L} =
	\lambda_1 \mathcal{L}{_{GAN}^{img}} +
	\lambda_2 \mathcal{L}{_1^{img}} +
	\lambda_3 \mathcal{L}{_P^{img}} +
	\lambda_4 \mathcal{L}{_{GAN}^{obj}} +
	\lambda_5 \mathcal{L}{_{AC}^{obj}} + 
	\lambda_6 \mathcal{L} {_{GAN} ^{phr}} +
	\lambda_7 \mathcal{L}_{DAMSM}^{phr} +
	\lambda_8 \mathcal{L}_{box}
\end{split}
\end{equation}
where $\lambda_i$ are the hyper parameters balancing various losses.

\section{Experiments}
\subsection{Experiment Settings}
\label{sec:experiment}
\noindent \textbf{Dataset.}
We use the COCO (CC-BY 4.0 license) 2017 dataset~\cite{10.1007/978-3-319-10602-1_48} and COCO-Stuff 2017 dataset~\cite{Caesar_2018_CVPR} to evaluate our method. These datasets contain 118K train and 5K validation images with instance-wise annotations and 5 text descriptions for each image over 80 \textit{thing} categories and 92 \textit{stuff} categories.
As a pre-processing step, we keep images containing 3$\sim$8 objects and drop objects covering less than 2\% area of the image. We split the COCO 2017 and COCO-Stuff 2017 validation set into our own validation and test sets, leaving us with 74121 train, 1024 validation, and 2048 test images.

\begin{table}[t]
\parbox{.55\linewidth}{
\begin{center}
\small
\caption{The quantitative performance on COCO-Stuff dataset.
The superscript $0, 1$ indicate inference with predicted boxes and gt-boxes respectively.
The results marked with $\star$ are trained by ourselves and marked with $\dagger$ are those reported in the original papers.}
\begin{tabular}[t]{l|c|c|c}
\hline
{Methods} &{IS} &{FID} & {Resolution}\\ \hline

{GT-Image} &$16.3 \pm 0.3$ & - & $64 \times 64$ \\
{GT-Image} &$26.14 \pm 0.4$ & - & $128 \times 128$ \\
{GT-Image} &$34.64 \pm 0.7$ & - & $256 \times 256$ \\ \hline

{MOC-GAN$^{0}$} &$\mathbf{28.96 \pm 0.4}$ &$\mathbf{25.89}$ & $256 \times 256$ \\
{MOC-GAN$^{1}$} &$31.72 \pm 0.5$ &$24.11$ & $256 \times 256$ \\
{Sg2Im $\star$} & $7.30 \pm 0.2$ & $43.82$ & $64 \times 64$ \\ 
{AttnGAN $\star$} & $24.41 \pm 0.4$ & $29.14$ & $256 \times 256$ \\ \hline

{Sg2Im~\cite{johnson2018image}$\dagger$} &$6.7 \pm 0.1$ & - & $64 \times 64$ \\
{AttnGAN~\cite{xu2017attngan}$\dagger$} &$23.79 \pm 0.32$ &$28.76$ & $256 \times 256$ \\
{StackGAN~\cite{han2017stackgan}$\dagger$} &$7.88 \pm 0.07$ & - & $64 \times 64$ \\
{PasteGAN~\cite{li2019pastegan}$\dagger$} &$9.1 \pm 0.2$ &$50.94$ & $64 \times 64$ \\
{Obj-GAN~\cite{li2019objgan}$\dagger$} &$27.37 \pm 0.22$ &$25.85$ & $256 \times 256$ \\
{Infer~\cite{Hong2018InferringSL}$^{0}\dagger$} &$11.46 \pm 0.09$ & - & $128 \times 128$ \\ \hline
\end{tabular}
\label{tab:quantative}
\end{center}
\vspace{-5mm}
}
\hfill
\parbox{.43\linewidth}{
\small
\begin{center}
\vspace{-2.3mm}
\caption{Ablation Study using IS and FID on COCO dataset. We use $\star$ to indicate that the model are trained under our data setting. $V$ means \textit{version}, "$-$" represents removing the corresponding feature maps in HFA. $low$ represents the low-resolution ($64 \times 64$) version.}
\label{tab:ablation_study}
\begin{tabular}{l|c|c}
\hline
	Methods  &  IS  &  FID    \\
	\hline
	Sg2Im $\star$ & $7.30 \pm 0.2$ & $43.82$ \\ \hline
	$V_\text{base}^{low}$ & $6.90 \pm 0.2$ & $43.63$ \\
	$V_\text{full}^{low}$ & $12.25 \pm 0.3$ & $38.11$ \\
	$V_\text{full}^{low} - L^\text{ig}$ & $10.07 \pm 0.4$ & $39.26$ \\
	$V_\text{base}$ & $11.20 \pm 0.2$ & $39.02$ \\
	$V_\text{full} - L^\text{ig} - c^\text{phr}_i$ & $ 17.22 \pm 0.6$ & $33.10$ \\
	$V_\text{full} - L^\text{phr}_i$ & $ 25.53 \pm 0.4$ & $28.37$ \\
	$V_\text{full} - c^\text{phr}_i$ & $ 19.84 \pm 0.3$ & $32.35$ \\
	\hline
	$V_\text{full}$ & $ 28.96 \pm 0.4$ & $25.89$ \\ \hline
\end{tabular}
\end{center}
}
\end{table}

\noindent \textbf{Implementation Details.}
We train all models using Adam~\cite{dpk2015adam} with a learning rate 5e-4 and batch size of 32 for 200,000 iterations, and the whole training process takes about 3$\sim$4 days on 8 GeForce 1080 Ti. The $\lambda_1 \sim \lambda_{8}$ are tuned heuristically and set to 1, 1, 0.5, 1, 0.1, 0.5, 5 and 10 respectively. ReLU is applied for graph convolution and Image Decoder while discriminators use LeakyReLU activation. More network structure details are introduced in the supplementary material.

\subsection{Metrics}
\noindent \textbf{Inception Score.}
We choose the Inception Score (IS)~\cite{NIPS2016_6125} by applying pre-trained Inception-v3 model~\cite{sze2016incept3} to compute the quality and diversity of the synthesized images, through calculating the KL divergence between the conditional class distribution and the marginal class distribution.

\noindent \textbf{Fr\'{e}chet Inception Distance.}
Different from IS that reflects the diversity of the entire set of synthesized images, the distance between the generated data distribution and the real data distribution is measured by FID~\cite{Heusel2017GANsTB}, which calculates the Wasserstein-2 distance between a pair of images in the feature space of Inception-v3 network.

\subsection{Quantitative and Qualitative Results}
The two most relevant scene image generation models are compared with our work: Sg2Im~\cite{johnson2018image}, the first work to generate images from scene graphs, and AttnGAN~\cite{xu2017attngan}, one of the SOTA method to generate images from captions. We use the codes released by them to train new models under our data splits for a fair comparison.
As shown in Table~\ref{tab:quantative}, MOC-GAN$^0$ outperforms AttnGAN $4.55$ and $3.25$, outperforms Sg2Im $21.66$ and $17.93$ on IS and FID respectively.
We list the results reported in the original papers of several previous methods, which could prove that MOC-GAN has achieved state-of-the-art quantitative performance.
Besides, the scores on ground-truth images are also reported for comparing with MOC-GAN$^1$, which shows a pretty small gap between them.

The $1$st and $6$th columns in Figure~\ref{fig:qualitative} (a) and (b) show the qualitative comparisons between AttnGAN and our complete MOC-GAN. It's clear that ours respect the relationships with each other while AttGAN contains more unknown artifacts.
It also illustrates the difficulty for caption-based models to extract interactive and spatial information between objects.
Additionally, the pizza image of ours shows much more details compared to AttnGAN, though AttnGAN has a natural advantage in generating images filled with only one object.

To evaluate the performance between Sg2Im and MOC-GAN for the same resolution, we only generate $64\times64$ resolution images and denote this version as $V_\text{full}^{low}$.
In Figure~\ref{fig:qualitative} (a) and (b), the $2$nd and $4$th columns are the results of Sg2Im and $V_\text{full}^{low}$. We can find more fine-grained and textual features from the whole images as well as the objects' appearance. For instance, the horses generated by Sg2Im are confused while ours show better appearance details and relative location. And $V_\text{full}^{low}$ also outperforms Sg2Im $4.95$ on IS and $5.71$ on FID shown in Table~\ref{tab:ablation_study}.

\subsection{Ablation Study and Qualitative Comparison}
\begin{figure*}
\begin{center}
\includegraphics[width=1.\linewidth]{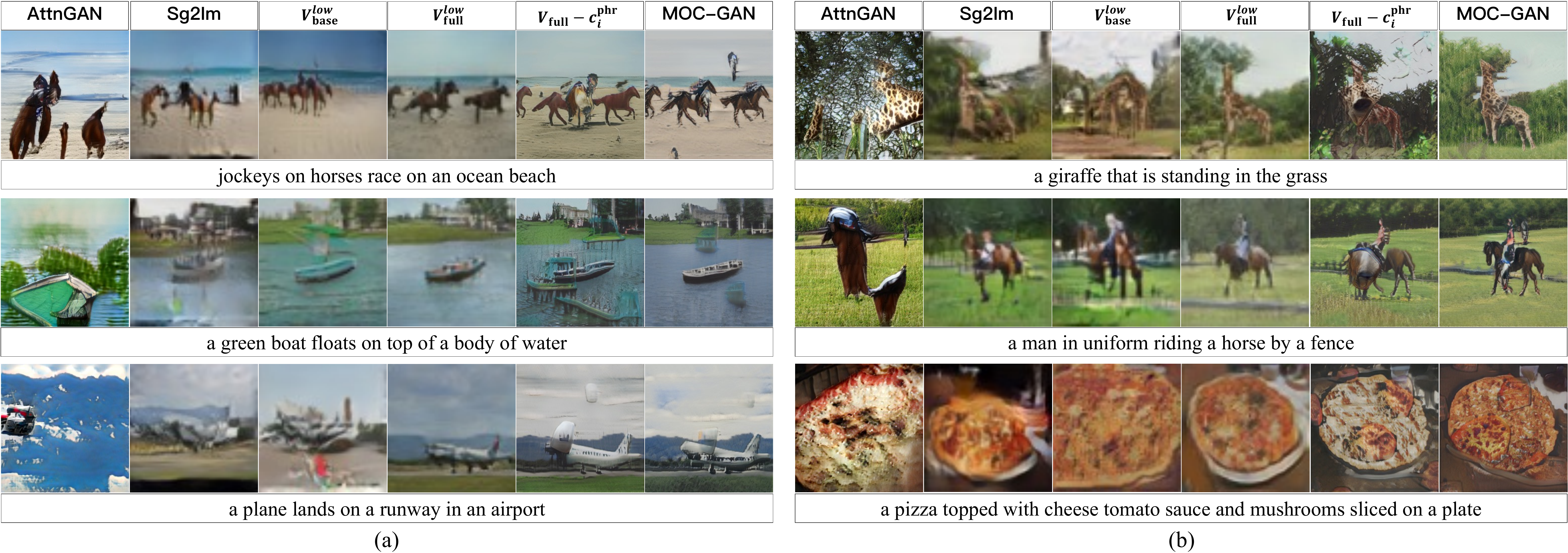}
\end{center}
\vspace{-5mm}
\caption{Examples of generated images on COCO dataset.
The result images of Sg2Im, $V_\text{base}^{low}$ and $V_\text{full}^{low}$ are all bilinearly upsampled to $256 \times 256$ from $64 \times 64$ resolution for better visualization.}
\label{fig:qualitative}
\vspace{-3mm}
\end{figure*}

To make a comparison for better verifying our proposed modules, we design a baseline $V_{\text{base}}$ taking input as both object labels and caption to generate images.
We keep the same pre-trained TE to extract words features $e$, and follow object-driven attention from Obj-GAN~\cite{li2019objgan} to encode the attribute information from $e$ for each object label. Specifically, the inputs for calculating the attention weight are the Glove embedding of object labels and words in the caption.
Then, we concatenate the object label embedding and its attribute encoding to get an integral representation feature for each object.
Next, the same structure BRN is utilized to process this integral feature to predict bounding box $b_i$. Then, we fill these features in their corresponding locations $b$ to form a layout map.
We could optionally control the resolution of layout map to generate $64 \times 64$ and $256 \times 256$ images, denoted as $V_\text{base}^{low}$ and $V_\text{base}$.
The unrelated loss weights are all disabled during training.
In Figure~\ref{fig:qualitative}, we visualize the generated images of previous methods and the ablated versions of our method. And the performance changes of essential components are shown in Table~\ref{tab:ablation_study}.

\textbf{$\boldsymbol{V_\text{base}^{low}}$ \textit{vs} $\boldsymbol{V_\text{full}^{low}}$.}
As shown in $3$rd and $4$th columns of Figure~\ref{fig:qualitative}, the results verify our intuition that $V_{\text{full}}^{low}$ outperforms $V_{\text{base}}^{low}$ to capture the spatial and interactive information, as well as the clear and realistic image quality, \eg, the boat generated by $V_\text{base}^{low}$ adheres to the riverbank with unreal relation, the plane generated by $V_\text{full}^{low}$ \textit{lands on a runway} rather than \textit{flies in the air.} Concretely, $V_{\text{base}}^{low}$ struggles on COCO dataset and yields only $6.9$ on IS and $43.63$ on FID, which are far less than $12.25$ and $38.11$ of $V_\text{full}^{low}$, even less than  $10.07$ and $39.26$ of $V_\text{full}^{low} - L^\text{ig}$. It indicates that the implicit graph structure processed with graph-convolution-based IGE could capture the spatial and interactive relationships, as well as the semantic attributes.

\noindent \textbf{Phrase-wise DAMSM.}
We evaluate the proposed phrase-wise DAMSM through removing phrase-context feature map $c^\text{phr}$ as $V_\text{full} - c^\text{phr}_i$, and show the results in $5$th column of Figure~\ref{fig:qualitative}. Compared to $6$th column, without the supervision of phrase-wise DAMSM, the generated phrase patches in $5$th column show disordered relationships, appearing many unknown phrase artifacts, these couple of objects may be fuzzy, \eg, the head of the plane tries to touch the sky, many boat-water pairs appear. In addition, compared to the $4$th column, it indicates the difficulties of high-resolution image generation. And this attention module could help the supervision with a coarse-to-fine process.

\section{Conclusions}
In this paper, we have proposed a novel cascaded attentive method for processing the object labels and captions to generate realistic images.
We propose a novel phrase-driven attention mechanism to infer the implicit relationships between any two objects, and a phrase-wise DAMSM to connect the graph and image semantic space, which supervise fin-grained phrase-level consistency to refine the generation process.
Compared to leading text-to-image generation algorithms which generate images from scene graphs or captions, our method implicitly represents the interactions between objects while maintaining high image quality.
A VGG-based phrase-wise discriminator is leveraged to provide rich phrase-wise discrimination signals on whether the synthesized object pairs match the high-level interactions between them.
Quantitive results, qualitative results and ablation study demonstrate the effectiveness and outstanding performance of our MOC-GAN.

\section*{Limitations}
In this section, we mainly introduce the limitations of our proposed method. Like other methods for generating images with multiple objects and complex interactive relationships, our model suffers from low quality and details loss due to the commonly known data bias and limit annotation issues, so that may produce very unreasonable and fake results. In addition, all experiments are conducted on COCO dataset with caption annotations and object labels, the generalization ability of our model needs further verification. And our model couldn't work when either of the two inputs doesn't exist.

\section*{Broader Impact}
\label{sec:impact}
MOC-GAN has the following positive impacts on society: (1) MOC-GAN could be used as a flexible text-to-image tool to generate images with complex objects and descriptions, which helps everyone to be an artist; (2) MOC-GAN could promote the development of scene image generation and even video generation in the future, maybe we could generate videos or films through only writing a script; (3) MOC-GAN may inspire the research of scene understanding aspects such as visual relation detection and image captioning. However, current research suffers from the low-quality details generation for a specific object from an image with multiple objects and complex interactions, we are still worried about the possibilities of generating human individuals full of face details like the real world in these complex interactive images, \eg, combine our MOC-GAN with other single-object generative models. That would cause some unnecessary troubles through fake or non-existent scenarios generation. On the other hand, researchers could be encouraged to develop more powerful models to learn the scene understandings and distinguish the authenticity of the generative scenarios for preventing potential security issues.

\small

\newpage
\appendix
\section{Text Encoder}
 We leverage a bi-directional LSTM based Text Encoder~(TE) to process the caption information following previous AttnGAN. In the bi-directional LSTM, each word corresponds to two hidden states, one for each direction. Thus, we concatenate its two hidden states to represent the semantic meaning of a word. The feature matrix of all words is indicated by $e \in \mathbb{R}^{D_w \times T_w}$. Its $i$-th column $e_i$ is the feature vector for the $i$-th word and $T_w$ is the number of words in a caption. Meanwhile, the last hidden states of the bi-directional LSTM are concatenated to be the global caption vector, denoted by $\overline{e} \in \mathbb{R} ^ {D_w}$.

The training of TE is associated with a Inception-V3 based Image Encoder, aiming to learn the word-patch similarity by decreasing the original DAMSM loss proposed in AttnGAN. We fix the parameters and utilize this pre-trained TE to process the captions in the following training procedures of Implicit Relation Estimator~(IRE) and phrase-wise DAMSM.

Note that the global caption vector $\overline{e}$ and this Image Encoder are deprecated in the subsequent model trainings. And please refer to Table~\ref{tab:notation} for notation details.

\begin{table}[hb]
\begin{center}
\caption{Notation.}
\label{tab:notation}
\begin{tabular}{c|c|l}
	\hline
	Variable  &  Type  &  Definition    \\
	\hline
	$o$ & Tensor & The embedding of object labels \\
	$v^o$ & Tensor & The feature vector of objects \\
	$v^r$ & Tensor & The feature vector of relations \\
	$v^\text{ir}$ & Tensor & Implicit relations \\
	$u$ & Tensor & The feature vector of phrases \\
	$\overline{u}$ & Tensor & The global feature of a whole implicit graph \\
	$e$ & Tensor & The feature vector of words \\
	$e^g$ & Tensor & Feature vector of words encoded by GloVe \\
	$\overline{e}$ & Tensor & The global feature vector of a whole caption \\
	$D_p$ & Scalar & Dimension of phrase feature vector \\
	$D_w$ & Scalar & Dimension of word feature vector \\
	$T_p$ & Scalar & Number of phrases in an implicit graph \\
	$T_w$ & Scalar & Number of words in a caption \\
	\hline
\end{tabular}
\end{center}
\end{table}

\section{Implicit Relation Estimator}
In this section, we mainly introduce the detailed architecture of our IRE and shown in Table~\ref{tab:ire}, we first adopt GloVe model to obtain the feature vector for every object label and word. By concatenating the subject $o_i^g \in \mathbb{R}^{50}$ and $o_k^g \in \mathbb{R}^{50}$ with a random sampled noise vector $z \in \mathbb{R}^{50}$, we could get a pseudo phrase feature vector. Then, two fully connected layers are utilized to refine this pseudo phrase vector and reduce the feature dimension, which produce a new phrase vector $q_j \in \mathbb{R}^{50}$ as mentioned in the main paper. At the similar space with GloVe semantics, we could calculate the attention weight through multiplying $q$ and $e^g$ as Equation~(2) in the main paper. Finally, the implicit relation $v^\text{ir}_j \in \mathbb{R}^{128}$ is formed as the weighted sum of words features $e$. The implicit graph could be formed through connecting any two objects with their corresponding implicit relation.

\begin{table}
\setlength{\tabcolsep}{6pt}
\begin{center}
\caption{The architecture of our Implicit Relation Estimator. The input is from (1) to (5). We mark the output feature as bold.}
\begin{tabular}[t]{|c|c|c|c|}
\hline
{Index} &{Input} &{Operation} & {Output Shape}\\ \hline
(1) & - & Glove subject vector $o^g_i$ & $50 $ \\
(2) & - & Noise vector $z$ & $50 $ \\
(3) & - & Glove object vector $o^g_k$ & $50 $ \\
(4) & - & Glove word features $e^g$ & $T_w \times 50 $ \\
(5) & - & Word features $e$ & $T_w \times 256 $ \\
(6) & (1)(2)(3) & Concatenate & $150 $ \\
(7) & (6) & Linear$(150 \rightarrow 300 )$ & $300 $ \\
(8) & (7) & ReLU & $ 300 $ \\
(9) & (8) & Linear$(300 \rightarrow 50)$ & $50$ \\
(10) & (9)(4) & Attention weight & $T_w$ \\
(11) & (5) & Linear$(256 \rightarrow 128)$ & $128$ \\
(12) & (11)(10) & Attention output & \bm{$128$} \\
\hline
\end{tabular}
\label{tab:ire}
\end{center}
\end{table}

\section{Phrase-wise DAMSM}
The phrase-wise DAMSM learns two neural networks that map sub-regions of the image and phrases of the implicit graph to a common semantic space, thus measures the image-graph similarity at the phrase level to compute a fine-grained loss for image generation. And the training procedure of phrase-wise DAMSM involves three neural networks: IRE to form the implicit graph, Implicit Graph Encoder~(IGE) to extract extract local phrase features and global graph features, and Image Encoder~(IE) for extracting image visual features. We first introduce the structure details of IGE and IE, and then describe how to calculate the image-graph similarity.

\subsection{Implicit Graph Encoder}
Our IGE is composed of three graph convolutional layers and two multi layer perceptrons (MLP). 
Firstly, the first graph convolutional layer takes as input the original implicit graph by concatenating any two object embedding $o_i$, $o_k$ with implicit relation vector $v^\text{ir}_j$.
After the processing of three graph convolutional layers, the spatial and interactive information are flowing along the edge and we deconcatenate the phrase feature to obtain the new object feature $v^o$ and relation feature $v^r$.
In addition, we adopt two MLPs to respectively further refine the phrase features and obtain two groups of features. 
One is a feature matrix of all phrases indicated by $u \in \mathbb{R} ^ {T_p \times D_p}$. Its $i$-th column $u_i$ is the feature vector for the $i$-th phrase. The other is a global graph vector $\overline{u} \in \mathbb{R}^{D_p}$, which is the fused phrases feature through average accumulation. $T_p$ is the number of phrases and $D_p$ is $128$ in our experiments. 
Table~\ref{tab:ige} shows the network operation in details.

\begin{table}
\setlength{\tabcolsep}{6pt}
\begin{center}
\caption{The architecture of our Implicit Graph Encoder. The input is a concatenation of object embedding and implicit relation vector in the form of (subject, predicate, object). All graph convolution layers have ReLU activation. $n$ is the number of objects. We mark the output feature as bold.}
\begin{tabular}[t]{|c|c|c|c|}
\hline
{Index} &{Input} &{Operation} & {Output Shape}\\ \hline
(1) & - & Subject embedding & $ n \times 128 $ \\
(2) & - & Implicit relation & $ T_p \times 128 $ \\
(3) & - & Object embedding & $ n \times 128 $ \\
(4) & (1)(2)(3) & Concatenate & $ T_p \times 384 $ \\
(5) & (4) & Graph conv layer & $ T_p \times 384 $ \\
(6) & (5) & Graph conv layer & $ T_p \times 384 $ \\
(7) & (6) & Graph conv layer & $ T_p \times 384 $ \\
(8) & (7) & Deconcatenate & $\bm{n \times 128, T_p \times 128}$ \\
(9) & (7) & Linear$(384 \rightarrow 768)$ & $ T_p \times 768 $ \\
(10) & (9) & ReLU & $ T_p \times 768 $ \\
(11) & (10) & Linear$(768 \rightarrow 384)$ & \bm{$ T_p \times 384 $} \\
(12) & (7) & Linear$(384 \rightarrow 768)$ & $ T_p \times 768 $ \\ 
(13) & (12) & ReLU & $ T_p \times 768 $ \\
(14) & (13) & Linear$(768 \rightarrow 384)$ & $ T_p \times 384 $ \\
(15) & (14) & Average Calculation & \bm{$384$} \\  \hline
\end{tabular}
\label{tab:ige}
\end{center}
\end{table}
 
\subsection{Image Encoder}
Inception-v3 model has been widely used as a backbone network to extract visual features, so we keep the same settings as previous AttnGAN to extract visual information for each image.
Specifically, we first rescale the input image and randomly crop it to be $ 299 \times 299 $ pixels to prevent over fitting and maintain generality. And then, we extract the local feature matrix $m \in \mathbb{R} ^ {768 \times 289} $ reshaped from $ 768 \times 17 \times 17 $ feature maps (from the ``\textit{mixed\_6e}" layer).
Simultaneously, the global feature vector $\overline{m} \in \mathbb{R}^{2048}$ is extracted from the last average pooling layer.
The additional 2 learning fully connected layers are used for converting these visual features to the semantic space, so we will get 2 visual features
\begin{equation}
	f = W m, \ \overline{f} = \overline{W}   \  \overline{m}
\end{equation}
where $f \in \mathbb{R} ^{128 \times 289} $ and $f_i$ is the visual feature vector for the $i$-th sub-region of the image; and $\overline{f} \in \mathbb{R} ^ {128}$ is the global vector for the whole image. During training, all original layers' parameters of Inception-v3 are fixed.
The detailed process is shown in Table~\ref{tab:ie}.

\begin{table}
\setlength{\tabcolsep}{6pt}
\begin{center}
\caption{The architecture of our Image Encoder. The input is a resized image. We mark the output feature as bold.}
\begin{tabular}[t]{|c|c|c|c|}
\hline
{Index} &{Input} &{Operation} & {Output Shape}\\ \hline
(1) & - & Image & $ 3 \times 299 \times 299 $ \\
(2) & (1) & Conv & $ 32 \times 149 \times 149 $ \\
(3) & (2) & Conv & $ 32 \times 147 \times 147 $ \\
(4) & (3) & Conv & $ 64 \times 147 \times 147 $ \\
(5) & (4) & Max Pooling & $ 64 \times 73 \times 73 $ \\
(6) & (5) & Conv & $ 80 \times 73 \times 73 $ \\
(7) & (6) & Conv & $ 192 \times 71 \times 71 $ \\ 
(8) & (7) & Max Pooling & $ 192 \times 35 \times 35 $ \\
(9) & (8) & Mixed\_5b & $ 256 \times 35 \times 35 $ \\
(10) & (9) & Mixed\_5c & $ 288 \times 35 \times 35 $ \\
(11) & (10) & Mixed\_5d & $ 288 \times 35 \times 35 $ \\ 
(12) & (11) & Mixed\_6a & $ 768 \times 17 \times 17 $ \\ 
(13) & (12) & Mixed\_6b & $ 768 \times 17 \times 17 $ \\ 
(14) & (13) & Mixed\_6c & $ 768 \times 17 \times 17 $ \\ 
(15) & (14) & Mixed\_6d & $ 768 \times 17 \times 17 $ \\ 
(16) & (15) & Mixed\_6e & $ 768 \times 17 \times 17 $ \\ 
(17) & (16) & Mixed\_7a & $ 1280 \times 8 \times 8 $ \\ 
(18) & (17) & Mixed\_7b & $ 2048 \times 8 \times 8 $ \\ 
(19) & (18) & Mixed\_7c & $ 2048 \times 8 \times 8 $ \\ 
(20) & (19) & Mixed\_7c & $ 2048 \times 1 \times 1 $ \\ 
(21) & (16) & Linear$(768 \rightarrow 128 )$ & $ 128 \times 17 \times 17 $ \\ 
(22) & (21) & Reshape & \bm{$128 \times 289 $} \\ 
(23) & (19) & Linear$(2048 \rightarrow 128 )$ & \bm{$128$} \\ 

\hline
\end{tabular}
\label{tab:ie}
\end{center}
\end{table}

\subsection{Image-Graph Matching Score}
The image-graph matching score is designed to measure the matching of an image-graph pair based on an attention model between the image and the implicit graph.

The similarity matrix for all possible pairs of phrases in the implicit graph and sub-regions in the image is calculated by 

\begin{equation}
	s = u^T f \ ,
\end{equation}

where $s \in \mathbb{R}^{T_p \times 289} $ and $s_{i,j}$ is the dot-product similarity between the $i$-th phrase of the implicit graph and the $j$-th sub-region of the image.
Each column is the similarity between this sub-region and every phrase in this graph. Each row is the similarity between this phrase and every sub-regions in this image.
Then we normalize the similarity between a sub-region and every phrase as follow, 

\begin{equation}
	\overline{s}_{i, j} = \frac{\exp{(s_{i,j}})}{{\sum_{k=0}^{T_{p}-1}}\exp{(s_{k, i})}}.
\end{equation}

An attention model is used to compute a region-context vector for each phrase (query). The region-context vector $c_i$ is a dynamic representation of the image's sub-regions related to the $i$-th phrase of the implicit graph, actually the relevance between a phrase and an whole image. It is computed by

\begin{equation}
	c_i = \sum ^{288}_{j=0}{\alpha _j} u_j, \
	\text{where} \
	\alpha _j = \frac{\exp({\gamma _1 \overline{s}_{i,j}})}{\sum _{k=0}^{288} \exp{({\gamma _1 \overline{s}_{i,k}})}} .
\end{equation}

Here, $\gamma _1$ is a factor. And that we do softmax operation in this dimension means the relevance between $i$-th phrase and all the sub-regions.

Finally, the relevance between the $i$-th phrase and the image is measured by the cosine similarity between $c_i$ and $u_i$, \ie, $R(c_i, u_i) = (c^T_i u_i )/(||c_i ||||u_i ||)$. The image-graph matching score between the entire image (X) and the whole graph (Y) is defined as
\begin{equation}
	R(X, Y) = \log(\sum ^{T_p-1} _{i=1} \exp{(\gamma _2 R(c_i, u_i))})^{\frac{1}{\gamma _2}}
\end{equation}

where $\gamma _2$ is a factor that determines how much to magnify the importance of the most relevant phrase-to-region-context pair. When $\gamma _2 \rightarrow \infty$, $R(X, Y)$ approximates to $\max^{T_p-1}_{i=1}R(c_i, u_i)$.

\subsection{Phrase-wise DAMSM Loss}
The phrase-wise DAMSM loss is designed to learn the matching between each phrase and each corresponding image patch. For a batch of image-graph pairs $\{ (X_i, Y_i) \}^M _{i=1}$, the posterior probability of graph $Y_i$ being matching with image $X_i$ is computed as
\begin{equation}
	P(Y_i | X_i) = \frac{\exp{(\gamma _3 R(X_i, Y_i))}}{\sum ^{M} _{j=1} \exp{(\gamma _3 R(X_i, Y_j))}}
\end{equation}
where $\gamma _3$ is a smoothing factor. In this batch of graphs, only $Y_i$ matches the image $X_i$, and treat all other $M-1$ graphs as mismatching pairs. The loss function could be defined as the negative log posterior probability that the images are matched with their corresponding implicit graphs (ground truth), \ie,
\begin{equation}
	\mathcal{L} ^p_1 = - \sum ^{M} _{i=1} \log{P(Y_i | X_i)},
\end{equation}

Meanwhile,
\begin{equation}
	\mathcal{L} ^p_2 = - \sum ^{M} _{i=1} \log{P(X_i | Y_i)},
\end{equation}

where $P(X_i|Y_i) = \frac{\exp{(\gamma _3 R(X_i, Y_i))}}{\sum ^{M} _{j=1} \exp{(\gamma _3 R(X_j, Y_i))}} $ is the posterior
probability that graph $Y_i$ is matched with its corresponding image $X_i$. Similarly, we can obtain loss functions $\mathcal{L}^g_1$ and $\mathcal{L}^g_2$ using the global graph vector $\overline{u}$ and the global image vector $\overline{f}$.
$\lambda_i$ are set to xxx respectively.

Finally, the phrase-wise DAMSM loss could be defined as
\begin{equation}
	\mathcal{L}^{phr}_{DAMSM} = \mathcal{L}^{p}_{1} + \mathcal{L}^{p}_{2} + \mathcal{L}^{g}_{1} + \mathcal{L}^{g}_{2} \ .
\end{equation}

\section{Box Regression}
We predict bounding boxes for images using a Box Regression Network~(BRN). Given an object feature vector $v^o$ produced by the IGE. The output from the box regression network is a predicted bounding box for the object, parameterized as $(x_0, y_0, x_1, y_1)$ where $x_0, x_1$ are the left and right coordinates of the box; all box coordinates are normalized to be in the range $[0, 1]$. The architecture of the box regression network is shown in Table~\ref{tab:box}.

\begin{table}[hb]
\setlength{\tabcolsep}{6pt}
\begin{center}
\caption{The architecture of our box regression network.}
\begin{tabular}[t]{|c|c|c|c|}
\hline
{Index} &{Input} &{Operation} & {Output Shape}\\ \hline
(1) & - & Object vector $v^o$ & $128 $ \\
(2) & (1) & Linear$(128 \rightarrow 512 )$ & $512 $ \\
(3) & (2) & ReLU & $ 512 $ \\
(4) & (3) & Linear$(512 \rightarrow 4)$ & $\textbf{4}$ \\
\hline
\end{tabular}
\label{tab:box}
\end{center}
\end{table}

\section{Image Decoder}
We separate Cascaded Refinement Netowork~(CRN) into two parts: Refiner and Output Head, to finish the final generation process. Specifically, the Refiner takes as input the comprehensive semantic map and outputs a hidden feature map $h$ which could be used in the next refinement process, and the Output Head is responsible for translating $h$ to image $\hat I$.
A Refiner is composed of a series of cascaded refinement modules~(CRM), and the feature map is upsampled between consecutive modules to make the spatial resolution double.
The input to each refinement module is a channel wise concatenation of the comprehensive semantic map (downsampled to the input resolution of the module to satisfy the requirement of module) and the feature map output by the previous refinement module. The exact architecture of CRM is shown in Table~\ref{tab:crn}.
The Output Head is implemented with $1\times1$ convolutional layers to convert the feature maps to 3-channel images.
The whole details of Image Decoder is shown in Table~\ref{tab:crn}.
 
\begin{table}
\setlength{\tabcolsep}{6pt}
\begin{center}
\caption{The architecture of a CRM. Each CRM takes as input the comprehensive semantic feature map and the feature map output by last CRM. And $H_\text{out} = 2H_\text{in}$ to let the resolution double between two neighboring CRMs. LeakyReLU uses a negative slope of $0.2$.}
\begin{tabular}[t]{|c|c|c|c|}
\hline
{Index} &{Input} &{Operation} & {Output Shape}\\ \hline
(1) & - & Comprehensive semantic feature & $D \times H \times W$ \\
(2) & - & Input features & $D_\text{in} \times H_\text{in} \times W_\text{in}$ \\
(3) & (1) & Average Pooling & $D \times H_\text{out} \times W_\text{out}$ \\
(4) & (2) & Upsample & $D_\text{in} \times H_\text{out} \times W_\text{out}$ \\
(5) & (3)(4) & Concatenate & $(D+D_\text{in}) \times H_\text{out} \times W_\text{out}$ \\
(6) & (5) & Conv($3\times3, D+D_\text{in} \rightarrow D_\text{out}$) & $D_\text{out} \times H_\text{out} \times W_\text{out}$ \\
(7) & (6) & BN & $D_\text{out} \times H_\text{out} \times W_\text{out}$ \\
(8) & (7) & LeakyReLU & $D_\text{out} \times H_\text{out} \times W_\text{out}$ \\
(9) & (8) & Conv($3\times3, D_\text{out} \rightarrow D_\text{out}$) & $D_\text{out} \times H_\text{out} \times W_\text{out}$ \\
(10) & (9) & BN & $D_\text{out} \times H_\text{out} \times W_\text{out}$ \\
(11) & (10) & LeakyReLU & $D_\text{out} \times H_\text{out} \times W_\text{out}$ \\
\hline
\end{tabular}
\label{tab:crm}
\end{center}
\end{table}

\begin{table}
\setlength{\tabcolsep}{6pt}
\begin{center}
\caption{The architecture of our Image Decoder, composed of Refiner and Output Head. LeakyReLU uses a negative slope of $0.2$. We only show the dimension changes of feature maps at first stage.}
\begin{tabular}[t]{|c|c|c|c|}
\hline
{Index} &{Input} &{Operation} & {Output Shape}\\ \hline
\multicolumn{4}{|c|}{Refiner} \\ \hline
(1) & - & Comprehensive semantic feature & $256 \times 64 \times 64$ \\
(2) & (1) & CRM($2\times2, 256 \rightarrow 1024$) & $1024 \times 4 \times 4$ \\
(3) & (1)(2) & CRM($4\times4, 1280 \rightarrow 512$) & $512 \times 8 \times 8$ \\
(4) & (1)(3) & CRM($8\times8, 768 \rightarrow 256$) & $256 \times 16 \times 16$ \\
(5) & (1)(4) & CRM($16\times16, 512 \rightarrow 128$) & $128 \times 32 \times 32$ \\
(6) & (1)(5) & CRM($32\times32, 384 \rightarrow 64$) & $64 \times 64 \times 64$ \\ \hline
\multicolumn{4}{|c|}{Output Head} \\ \hline
(7) & (6) & Conv($3\times3, 64 \rightarrow 64$) & $64 \times 64 \times 64$ \\
(8) & (7) & LeakyReLU & $64 \times 64 \times 64$ \\
(9) & (8) & Conv($1\times1, 64 \rightarrow 3$) & $3 \times 64 \times 64$ \\
\hline
\end{tabular}
\label{tab:crn}
\end{center}
\end{table}

\section{Discriminator}
\subsection{Patch-wise Discriminator}
Our unconditional patch-wise discriminator takes as input a real or a fake image and classifies the input as real or fake. The architecture details are shown in Table~\ref{tab:disimgunc}.
And the implicit graph conditional patch-wise discriminator takes as input the global graph feature and a real or a fake image, and whether the input image is consistent with the implicit graph or not. The architecture details are shown in Table~\ref{tab:disimgcon}.
\begin{table}
\setlength{\tabcolsep}{6pt}
\begin{center}
\caption{The architecture of our unconditional patch-wise discriminator $D_\text{unc}^\text{pat}$. The input to this discriminator is either a real or fake image. $s2$ means the stride of convolution is $2$, and all convolutions use no padding. LeakyReLU uses a negative slope of $0.2$.  We show the feature dimension changes for $64 \times 64$ resolution images.}
\begin{tabular}[t]{|c|c|c|c|}
\hline
{Index} &{Input} &{Operation} & {Output Shape}\\ \hline
(1) & - & Image & $3 \times 64 \times 64$ \\
(2) & (1) & Conv($4\times4, 3 \rightarrow 64, s2$) & $64 \times 32 \times 32$ \\
(3) & (2) & BN & $64 \times 32 \times 32$ \\
(4) & (3) & LeakyReLU & $64 \times 32 \times 32$ \\
(5) & (4) & Conv($4\times4, 64 \rightarrow 128, s2$) & $128 \times 16 \times 16$ \\
(6) & (5) & BN & $128 \times 16 \times 16$ \\
(7) & (6) & LeakyReLU & $128 \times 16 \times 16$ \\
(8) & (7) & Conv($4\times4, 128 \rightarrow 256, s2$) & $256 \times 8 \times 8$ \\
(9) & (8) & Conv($1\times1, 256 \rightarrow 1$) & $1 \times 8 \times 8$ \\
\hline
\end{tabular}
\label{tab:disimgunc}
\end{center}
\end{table}

\begin{table}
\setlength{\tabcolsep}{6pt}
\begin{center}
\caption{The architecture of our implicit graph conditional patch-wise discriminator $D_\text{ig\_con}^\text{pat}$. The input to this discriminator is the global graph feature and either a real or a fake image. LeakyReLU uses a negative slope of $0.2$. We show the feature dimension changes for $64 \times 64$ resolution images.}
\begin{tabular}[t]{|c|c|c|c|}
\hline
{Index} &{Input} &{Operation} & {Output Shape}\\ \hline
(1) & - & Image & $3 \times 64 \times 64$ \\
(2) & - & Global graph feature & $128$ \\
(3) & (1) & Conv($4\times4, 3 \rightarrow 64, s2$) & $64 \times 32 \times 32$ \\
(4) & (3) & BN & $64 \times 32 \times 32$ \\
(5) & (4) & LeakyReLU & $64 \times 32 \times 32$ \\
(6) & (5) & Conv($4\times4, 64 \rightarrow 128, s2$) & $128 \times 16 \times 16$ \\
(7) & (6) & BN & $128 \times 16 \times 16$ \\
(8) & (7) & LeakyReLU & $128 \times 16 \times 16$ \\
(9) & (2) & Repeat & $128 \times 16 \times 16$ \\
(10) & (8)(9) & Concatenate & $256 \times 16 \times 16$ \\
(11) & (10) & Conv($4\times4, 256 \rightarrow 512, s2$) & $512 \times 8 \times 8$ \\
(12) & (11) & Conv($1\times1, 512 \rightarrow 1$) & $1 \times 8 \times 8$ \\
\hline
\end{tabular}
\label{tab:disimgcon}
\end{center}
\end{table}

\subsection{Object-wise Discriminator}
The object-wise discriminator $D^{obj}$ inputs object pixels in the real or generated images. They are all cropped using their bounding boxes and resized to the half of the image size using bilinear interpolation. The object-wise discriminator plays two roles: distinguish the objects as real or fake, classifier the category of the objects. The architecture details are shown in Table~\ref{tab:disobj}.
\begin{table}[hb]
\setlength{\tabcolsep}{6pt}
\begin{center}
\caption{The architecture of our object-wise discriminator $D^\text{obj}$ and $D_\text{ac}^\text{obj}$. $\mathcal{C}$ indicates the number of object categories. LeakyReLU uses a negative slope of $0.2$. We show the for $64 \times 64$ resolution images.}
\begin{tabular}[t]{|c|c|c|c|}
\hline
{Index} &{Input} &{Operation} & {Output Shape}\\ \hline
(1) & - & Object crop & $3 \times 32 \times 32$ \\
(2) & (1) & Conv($4\times4, 3 \rightarrow 64, s2$) & $64 \times 16 \times 16$ \\
(3) & (2) & BN & $64 \times 16 \times 16$ \\
(4) & (3) & LeakyReLU & $64 \times 16 \times 16$ \\
(5) & (4) & Conv($4\times4, 64 \rightarrow 128, s2$) & $128 \times 8 \times 8$ \\
(6) & (5) & BN & $128 \times 8 \times 8$ \\
(7) & (6) & LeakyReLU & $128 \times 8 \times 8$ \\
(8) & (7) & Conv($4\times4, 128 \rightarrow 256, s2$) & $256 \times 4 \times 4$ \\
(9) & (8) & Average Pooling & $256$ \\
(10) & (9) & Linear($256 \rightarrow 1024$) & $1024$ \\
(11) & (10) & Linear($1024 \rightarrow 1$) & $1$ \\
(12) & (10) & Linear($1024 \rightarrow \mathcal{C}$) & $\mathcal{C}$ \\
\hline
\end{tabular}
\label{tab:disobj}
\end{center}
\end{table}

\subsection{Phrase-wise Discriminator}
We adopt a pre-trained VGG19 network~(from the last max pooling layer) as a feature extraction backbone to extract the visual feature maps of \textit{subject}, \textit{predicate} and \textit{object}.
For unconditional phrase-wise discriminator, we concatenate these features to form the phrase feature and try to classifier the phrase as real or fake. The architecture details are shown in Table~\ref{tab:disphrunc}.
For conditional phrase-wise discriminator, we use implicit relationship $v^\text{ir}$ to replace the position of \textit{predicate} feature, and to determine whether these two objects are correlated with the relations attended from words features. The architecture details are shown in Table~\ref{tab:disphrcon}.
\begin{table}[hb]
\setlength{\tabcolsep}{6pt}
\begin{center}
\caption{The architecture of our unconditional phrase-wise discriminator $D_\text{unc}^\text{phr}$. LeakyReLU uses a negative slope of $0.2$. We show the for $256 \times 256$ resolution images.}
\begin{tabular}[t]{|c|c|c|c|}
\hline
{Index} &{Input} &{Operation} & {Output Shape}\\ \hline
(1) & - & Subject crop & $3 \times 128 \times 128$ \\
(2) & - & Predicate crop & $3 \times 128 \times 128$ \\
(3) & - & Object crop & $3 \times 128 \times 128$ \\
(4) & (1) & VGG19 & $512 \times 4 \times 4$ \\
(5) & (2) & VGG19 & $512 \times 4 \times 4$ \\
(6) & (3) & VGG19 & $512 \times 4 \times 4$ \\
(7) & (4)(5)(6) & Concatenate & $1536 \times 4 \times 4$ \\
(8) & (7) & Conv($3\times3, 1536 \rightarrow 512, s1$) & $512 \times 2 \times 2$ \\
(9) & (8) & BN & $512 \times 2 \times 2$ \\
(10) & (9) & LeakyReLU & $512 \times 2 \times 2$ \\
(11) & (10) & Conv($1\times1, 512 \rightarrow 1$) & $1 \times 2 \times 2$ \\
\hline
\end{tabular}
\label{tab:disphrunc}
\end{center}
\end{table}

\newpage

\begin{table}[htbp]
\setlength{\tabcolsep}{6pt}
\begin{center}
\caption{The architecture of our implicit relationship conditional phrase-wise discriminator $D_\text{con}^\text{phr}$. LeakyReLU uses a negative slope of $0.2$. We show the feature shape changes for $256 \times 256$ resolution images.}
\begin{tabular}[t]{|c|c|c|c|}
\hline
{Index} &{Input} &{Operation} & {Output Shape}\\ \hline
(1) & - & Subject crop & $3 \times 128 \times 128$ \\
(2) & - & Implicit relationship & $128$ \\
(3) & - & Object crop & $3 \times 128 \times 128$ \\
(4) & (1) & VGG19 & $512 \times 4 \times 4$ \\
(5) & (2) & Repeat & $128 \times 4 \times 4$ \\
(6) & (3) & VGG19 & $512 \times 4 \times 4$ \\
(7) & (4)(5)(6) & Concatenate & $1152 \times 4 \times 4$ \\
(8) & (7) & Conv($3\times3, 1152 \rightarrow 512, s1$) & $512 \times 2 \times 2$ \\
(9) & (8) & BN & $512 \times 2 \times 2$ \\
(10) & (9) & LeakyReLU & $512 \times 2 \times 2$ \\
(11) & (10) & Conv($1\times1, 512 \rightarrow 1$) & $1 \times 2 \times 2$ \\
\hline
\end{tabular}
\label{tab:disphrcon}
\end{center}
\end{table}

\begin{figure}
\centering
\includegraphics[width=1.\linewidth]{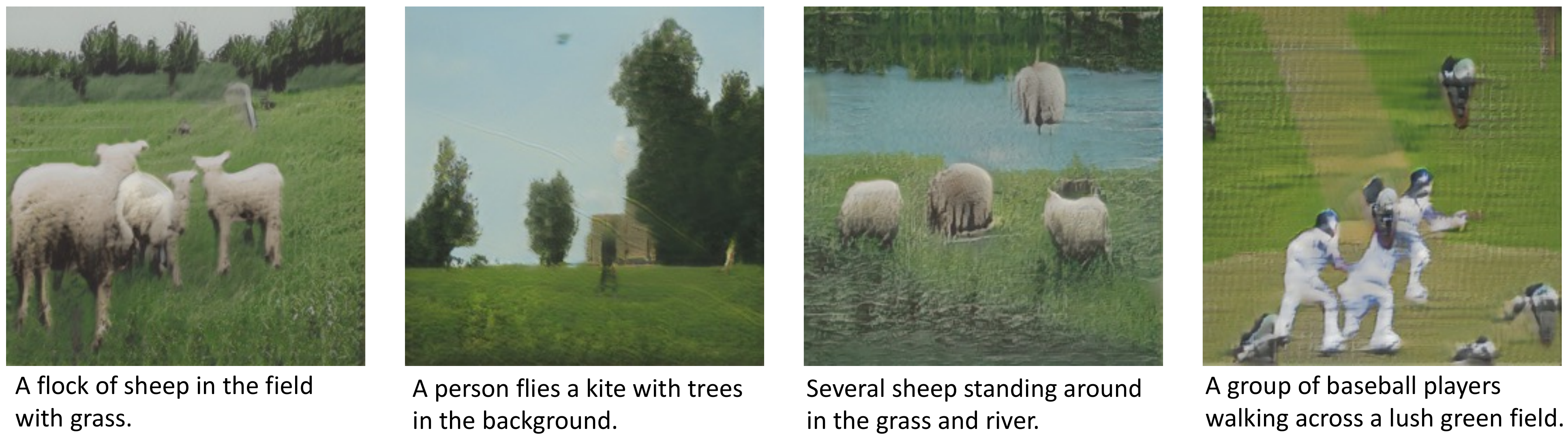}
\caption{Some generated examples of MOC-GAN.}
\label{fig:pic}
\end{figure}

\end{document}